\pdfoutput=1

\documentclass[11pt]{article}

\usepackage{enumitem}

\usepackage{acl}

\usepackage{times}

\usepackage{latexsym}

\usepackage{booktabs}

\usepackage[utf8]{inputenc}
\usepackage{graphicx}

\usepackage[T1]{fontenc}

\usepackage[utf8]{inputenc}

\usepackage{microtype}

\usepackage{inconsolata}

\usepackage{wrapfig}
\usepackage{float}
\usepackage{graphicx}
\usepackage{caption}
\usepackage{amsmath}
\usepackage{url}
\usepackage{cite}

\title{Assessing the Reliability and Validity of GPT-4 in Annotating Emotion Appraisal Ratings}

\author{Deniss Ruder \\ Institute of Computer Science \\  University of Tartu \\ \texttt{deniss.ruder@ut.ee}
         \And  Andero Uusberg \\ Institute of Psychology \\  University of Tartu \\ \texttt{andero.uusberg@ut.ee} \And Kairit Sirts \\ Institute of Computer Science \\  University of Tartu \\ \texttt{kairit.sirts@ut.ee}}

\begin{document}
\maketitle
\begin{abstract}
Appraisal theories suggest that emotions arise from subjective evaluations of events, referred to as appraisals. The taxonomy of appraisals is quite diverse, and they are usually given ratings on a Likert scale to be annotated in an experiencer-annotator or reader-annotator paradigm. This paper studies GPT-4 as a reader-annotator of 21 specific appraisal ratings in different prompt settings, aiming to evaluate and improve its performance compared to human annotators. We found that GPT-4 is an effective reader-annotator that performs close to or even slightly better than human annotators, and its results can be significantly improved by using a majority voting of five completions. GPT-4 also effectively predicts appraisal ratings and emotion labels using a single prompt, but adding instruction complexity results in poorer performance. We also found that longer event descriptions lead to more accurate annotations for both model and human annotator ratings. This work contributes to the growing usage of LLMs in psychology and the strategies for improving GPT-4 performance in annotating appraisals.

\end{abstract}

\section{Introduction}
\label{sec:intro}

According to appraisal theories, emotions emerge from the individual's subjective appraisals of significant events \citep{scherer2009dynamic}. Appraisals are the person's evaluations of what situations mean for their needs, goals, and other concerns \citep{moors2013appraisal}. Appraisals consist of values on abstract dimensions representing key aspects of a situation, such as how important, desirable, self-caused, certain, and controllable it is.  Appraisals orchestrate changes in other components of an emotional episode, including tendencies to act in some way (motivational component), visceral preparations for these actions (somatic component), facial and bodily expressions (motor component), and a conscious feeling (experiential component) \citep{moors2013appraisal}. Even as all of these components can influence each other, appraisals are often considered as pivotal for initiating and shaping the dynamic interactions that underlie a person's experience of emotion.

Despite their centrality in emotion theory, there have been only few attempts to apply NLP methods to automatically extracting appraisals from text \citep{hofmann2020appraisal,hofmann2021emotion,troiano-etal-2023-dimensional}. 
In those previous works, appraisals have been of interest as a component of emotions, with the goal of eventually predicting the emotion themselves \citep{hofmann2020appraisal,troiano-etal-2023-dimensional}. These early results are promising, but more research is needed to further improve the precision and robustness of appraisal prediction models.

Advancing the NLP research on appraisal requires suitable annotated datasets. The most accurate way to annotate appraisals is the so-called experiencer-annotator paradigm, where a person 
provides both a textual description of some emotional event and ratings on appraisal dimensions. However, the experiencer-annotator method can only be used on texts collected specifically for this purpose. In an alternative reader-annotator paradigm, the appraisal ratings are provided by another person reading the textual descriptions of emotional events. The reader-annotator method is more flexible, as it enables using existing datasets of emotional event descriptions, even those that have not been specifically collected for that purpose but rather have been generated spontaneously by people, for instance in social media.

Generating appraisal ratings with the reader-annotator procedure requires human labor that can become prohibitively expensive for large real-world datasets such as social media and blog posts. Therefore, we are interested whether this process can be automated. Previously, \citet{hofmann2021emotion} adopted a rule-based approach to assign appraisal labels to texts, given that the emotion label of the text was known. However, this approach requires the emotion labels which are often unavailable in real-world datasets. Also,  \citet{hofmann2021emotion} considered only six appraisals and represented them as binary variables. Thus, the deterministic rule-based approach is clearly not feasible with a larger number of appraisal dimensions assessed on Likert scales.

The goal of this work is to assess the suitability of Large Language Models (LLMs) to act as an annotator in the reader-annotator paradigm for emotional appraisals. Several previous works have shown the utility of using LLMs as a viable alternative for human annotators for labeling data for NLP tasks using both the proprietary GPT models \citep{ding-etal-2023-gpt} as well as open-source LLMs \citep{alizadeh2023open}. Other studies have found that using GPT-4 is generally  a more reliable annotator than crowdworkers for social science related text annotations \citep{gilardi2023chatgpt}, and it is better than vocabulary-based methods for annotating several psychological constructs \citep{rathje2024gpt}. Although several studies have assessed the ability of GPT-4 to detect discrete emotion labels \citep{niu2024text,lian2024gpt}, \citet{kocon2023chatgpt} found that GPT-4 was performing worse than fine-tuned classification models, especially on emotion prediction tasks. 

Our study focuses on adopting GPT-4 for annotating emotion appraisals in the reader-annotator paradigm to assess the viability of the generative LLM to act as an alternative to human reader-annotators. Few studies have explored the capabilities of GPT-4 and similar LLMs in this context. \citet{tak2023gpt} analyzed the emotional reasoning abilities of GPT models, finding that while they align well with human appraisals, they struggled with predicting emotion intensity and coping response. \citet{yongsatianchot2023investigating} studied how LLMs perceive emotions using appraisal theory, demonstrating that these models can effectively generate context-specific emotional appraisals. However, these studies focused more on the rationales of LLMs than appraisal rating predictions and used a limited list of appraisal dimensions. \citet{zhan-etal-2023-evaluating} evaluated annotating emotion appraisals by multiple LLMs, showing that they can produce human-like emotional appraisals. 
However, the dataset used in this research was relatively small and, more importantly, did not contain original experience-annotator ratings, which prevents comparison between the event experiencer and reader annotations.

For our study, we use the crowd-enVent dataset collected by \citet{troiano-etal-2023-dimensional}, which consists of crowd-sourced emotional event descriptions, supplied by the appraisal rating of both the experiencers and readers. We formulate research questions about the \emph{reliability} of GPT-4 in generating appraisal ratings (Q1) and the \emph{accuracy} of GPT-4 ratings compared to both experiencer-annotators and human reader-annotators (Q2). In addition, we test whether the application of \textit{a majority voting algorithm} and a model confidence tiebreaker improves accuracy (Q3). Also, we examine the impact of adding \emph{the emotion prediction} on the accuracy of the appraisal ratings (Q4). Finally, we study the impact of the event \emph{description length} on the accuracy of the ratings by both the GPT-4 and human reader-annotators (Q5).


\section{Research Questions}
\label {sec:questions}
We pose two main research questions to evaluate the reliability and accuracy of GPT-4 to act as an annotator for emotion appraisals in the reader-annotator paradigm. Also, we pose three additional research questions that study the impact of a majority voting algorithm, the prediction of emotions, and the length of the event description on the accuracy of the appraisals. This section outlines our research questions.

\paragraph{Q1: Is GPT-4 reliable in generating emotion appraisal ratings?}
Reliability is a prerequisite for validity. For the GPT-4 to act as a valid annotator of the emotion appraisals, the reliability of its ratings needs to be high in terms of good inter-annotator agreement between several independent GPT-4 runs. We assessed the reliability of the appraisal ratings in a subsample of 108 randomly sampled texts and measured the reliability using  Spearman correlation coefficients and root mean squared errors (RMSE). For GPT-4 to be considered reliable, the ratings of different runs should show at least the same level of agreement as different human reader-annotators.

\paragraph{Q2: How accurate are the GPT-4 appraisal ratings compared to the ratings given by the human reader-annotators?} 
To validate GPT-4 as an accurate annotator of emotion appraisals based on event descriptions, we examined the difference between the ratings assigned by GPT-4 and those assigned by experiencer-annotators. We compared these differences with the rating differences between experiencer-annotators and reader-annotators. For GPT-4 to be considered a valid emotion appraisal annotator, its ratings should remain at least close to those of the human reader-annotators ratings.

\paragraph{Q3: Can majority voting  with or without a confidence tiebreaker improve GPT-4 accuracy?}
Applying a majority voting algorithm on a human reader-annotators dataset greatly affected the accuracy of reader guesses \citep{troiano-etal-2023-dimensional}. The authors also noticed that a substantial number of votes required tiebreakers and proposed breaking the ties by assigning a higher weight to the annotators with stronger confidence. We assumed that majority voting might similarly
impact GPT-4 performance, and we can improve it with a model confidence tiebreaker.

\paragraph{Q4: Does adding an emotion prediction task to the prompt impact appraisal ratings accuracy?}
During the crowd-enVent dataset collection process, human reader-annotators were first asked to select an emotion that the event experiencer-annotator likely felt (anger, boredom, disgust, fear, guilt, joy, pride, relief, sadness, shame, surprise trust, no emotion) as well as to rate how confident they were about their chosen emotion (1-5) and the intensity of emotion (1-5). 
This procedure makes sense both intuitively and theoretically, as different emotions are expected to have different signature appraisal profiles \citep{moors2013appraisal}. Thus, knowing/guessing the emotion might help to predict/guess also the appraisal ratings more accurately.
We tested if reproducing this exact sequence of questions improves GPT-4 appraisal prediction accuracy. We also attempted to use the newly generated confidence and intensity ratings as majority voting tiebreakers.

\paragraph{Q5: How does the length of the event description affect the accuracy of the appraisal ratings for both the GPT-4 and the human reader-annotators?} 
The length of the description of the emotional event is one practical consideration in the reader-annotator paradigm. While the experiencer-annotator will have direct access to the emotion and the related appraisals, the reader, whether an LLM or a human, has to make inferences based on the written text only. It seems intuitive that longer descriptions would give rise to more accurate appraisal ratings. In this research question, we tested this intuition and sought to identify a minimum required length of text to make accurate predictions or guesses.

\section{Method}
\label {sec:method}

\subsection{Data}
\label{subsec:related}

We use the crowd-enVent data collected by \citet{troiano-etal-2023-dimensional}\footnote{Available in \url{https://github.com/sarnthil/crowd-enVent-modeling}} who explored the application of appraisal theories to the analysis of emotions in the text. They investigated whether human reader-annotators and an automatic RoBERTa-based text classifier can reproduce appraisal ratings of the original event experiencers and whether these appraisal ratings can assist in labeling emotions. 

The data collection process included two phases. During the first phase, the authors collected event descriptions, appraisal ratings, 
and the categorical emotion experienced in relation to the event from the experiencer-annotators. In total, 6600 event descriptions were collected via crowd-sourcing. In the second phase, a subset of 1200 event descriptions was subsequently annotated by five human crowd-sourced reader-annotators. In addition to guessing the appraisal ratings of the experiencers of the event, the reader-annotators also were asked to guess the emotion label related to the event, and rate their confidence in the 5-point Likert scale in that guess.
Thus, the final appraisal-annotated crowd-enVent corpus includes two datasets that we call experiencer-annotator and reader-annotator datasets, respectively. Both datasets are annotated with the following 21 appraisal dimensions on a 1-5 Likert scale.

\begin{enumerate}[leftmargin=*, label=\arabic*., itemsep=-0.25em]
\item\textbf{Suddenness}: The event was sudden or abrupt.
\item \textbf{Familiarity}: The event was familiar.
\item \textbf{Event Predictability}: I could have predicted the occurrence of the event.
\item \textbf{Pleasantness}: The event was pleasant.
\item \textbf{Unpleasantness}: The event was unpleasant.
\item \textbf{Goal Relevance}: I expected the event to have important consequences for me.
\item \textbf{Situational Responsibility}: The event was caused by chance, special circumstances, or natural forces.
\item \textbf{Own Responsibility}: The event was caused by my own behavior.
\item \textbf{Others’ Responsibility}: The event was caused by someone else's behavior.
\item \textbf{Anticipated Consequences}: I anticipated the consequences of the event.
\item \textbf{Goal Support}: I expected positive consequences for me.
\item \textbf{Urgency}: The event required an immediate response.
\item \textbf{Own Control}: I was able to influence what was going on during the event.
\item \textbf{Others’ Control}: Someone other than me was influencing what was going on.
\item \textbf{Situational Control}: The situation was the result of outside influences that no one had control.
\item \textbf{Accepted Consequences}: I accepted that I would easily live with the unavoidable consequences of the event.
\item \textbf{Internal Norms}: The event clashed with my standards and ideals.
\item \textbf{External Norms}: The actions that produced the event violated laws or socially accepted norms.
\item \textbf{Attention}: I had to pay attention to the situation.
\item \textbf{Not consider}: I tried to shut the situation out of my mind.
\item \textbf{Effort}: The situation required me a great deal of energy to deal with it.

\end{enumerate}

In addition, the experiencer-annotators and the reader-annotators had to choose one emotion they or the experiencer felt from a list of 13 emotions: anger, boredom, disgust, fear, guilt, joy, pride, relief, sadness, shame, surprise, trust and no emotion. 

\subsection{Model Setup}
\label{subsec:setup}

To generate appraisal ratings, we used the GPT-4\footnote{Model version: 0613} via the Azure OpenAI REST API service.\footnote{API version: 2023-07-01-preview} We adopted the default parameters of temperature 0.7 and top\_p 0.95, with both presence and frequency penalty 0. 
For Q1, we conducted 5 independent runs of a single prompt. For Q2, we performed only 1 run. For research questions Q2, Q3, and Q4, we set the number of completions GPT-4 parameter to 5. 

\subsection{Prompt}
\label{subsec:prompt}

Our prompt to the GPT-4 consists of three parts: the general context \textbf{C}, more detailed instructions \textbf{I} and the description of the event \textbf{D}.

\paragraph{Context:} The context part of the prompt gives an overall instruction to the model to act as an expert:
\\\\
\noindent
[C] \emph{``You are an expert in human psychology. You will read descriptions people have written about an event or situation eliciting an emotional reaction. You will see a series of questions about how people think about these events.''}

\paragraph{Instruction:} The instruction part of the prompt asks the model to give integer ratings to the listed questions and specifies the output format. We used two versions of the instruction. The first instruction, I1, used for Q1 and Q2, asks GPT-4 to assign ratings to appraisal dimensions. Q3 used the same instruction with one additional question, "How confident is the model about chosen ratings?". The second instruction, I2, used for Q4, asks for the prediction of the emotion label and the confidence and intensity ratings for the emotion.
\\\\
\noindent
[I1] \emph{Instruction: Based on event descriptions, assign integer ratings varying from 1 to 5 to the list of questions. \\
Desired format: plain values in this order: <list of appraisal titles> \\
List of questions: <list of appraisals> } 
\\\\
\noindent
[I2] \emph{Instruction: Based on descriptions, choose one emotion from the list of emotions and assign integer ratings varying from 1 to 5 to the list of questions. \\
Desired format: plain values in this order: emotion, confidence, intensity, <list of appraisal titles>  \\
List of emotions:  <list of emotion titles> \\
List of questions: How confident are you about your emotion?, "How intense do you think the emotion was?", <list of appraisals>} 

\paragraph{Description:}
For event descriptions, we used the version of the dataset in which explicit words of the target emotion were masked, to avoid the model using these words as superficial heuristics for predicting evaluation ratings. The same masked version of the text was used by \citet{troiano-etal-2023-dimensional} to collect the appraisal ratings by the annotators.
\\\\
\noindent
[D] \emph{``People get under my skin. Like for example if an entitled customer shows up at my work and demands to speak to my manager for a simple issue that I can resolve. This happens on almost a daily occurrence and it really makes me <masked>.''}

\section{Results}
\label{sec:results}

\subsection{Q1 Reliability of GPT-4}
\label{subsec:consistency}
We studied the Q1 on a subsample of the reader-annotator dataset.
We aimed for a subsample of at least 100 event descriptions stratified to match the emotion category distribution in the full dataset. In order to account for potential losses due to Azure policy and other model and format generation errors, we opted for taking a subsample of 9\% of the reader-annotator dataset. During appraisal rating generation, we encountered no policy errors, nor partial or empty outcomes, resulting in a subset on 108 event descriptions.
To assess GPT-4 reliability in generating appraisal ratings, we calculated Root Mean Squared Errors (RMSE) and Spearman correlation coefficients between all pairs of appraisal rating vectors from five GPT-4 runs\footnote{In total $\frac{n\cdot(n-1)}{2}=\frac{5\cdot4}{2}=10$ pairs.} for each appraisal dimension. We then averaged these metrics across all pairs. 

The average pairwise RMSE (APRMSE) values, macro-averaged over all appraisal dimensions, were very close to each other, ranging from 0.6 to 0.63 over the five runs, with a mean of 0.61 (see Table~\ref{tab:t1}. 
Across individual appraisal dimensions, the lowest (e.g., the most accurate) APRMSE value was observed for External Norms, while the model struggled the most with Anticipated Consequences, for which it obtained the highest (e.g., the least accurate) APRMSE score of 0.79.

Analyzing Spearman correlation coefficients, we observed strong and very strong correlations between responses from different GPT-4 runs. Mean Spearman coefficients ranged from 0.67 for Attention to 0.97 for Pleasantness, with a mean of 0.87 for the macro-average over all appraisal dimensions (\ref{tab:t1}). This compares favorably with common guidelines in psychometrics that consider test-retest correlations above 0.8 to indicate good reliability and above 0.90 to indicate excellent reliability.

\begin{table}[t]
    \centering
    \small
    \begin{tabular}{lcc} 
        \toprule
        Appraisal & APRMSE& $\rho$\\  
        \midrule  
        Suddenness & 0.69 & 0.89 \\   
        Familiarity & 0.77 & 0.84 \\ 
        Event predictability & 0.69 & 0.87 \\ 
        Pleasantness & 0.38 & 0.97 \\ 
        Unpleasantness & 0.42 & 0.96 \\ 
        Goal Relevance & 0.58 & 0.89 \\ 
        Situational Responsibility & 0.72 & 0.77 \\ 
        Own Responsibility & 0.62 & 0.91 \\ 
        Others' Responsibility & 0.62 & 0.91 \\ 
        Anticipated Consequences & 0.79 & 0.68 \\ 
        Goal Support& 0.59 & 0.93  \\ 
        Urgency & 0.69 & 0.86\\ 
        Own Control& 0.66 & 0.87\\ 
        Others' Control & 0.66 & 0.92 \\ 
        Situational Control & 0.76 & 0.81 \\ 
        Accepted Consequences & 0.58 & 0.84 \\ 
        Internal Norms & 0.57 & 0.93 \\ 
        External Norms & 0.37 & 0.92 \\ 
        Attention & 0.65 & 0.67 \\ 
        Not Consider & 0.51 & 0.94 \\ 
        Effort & 0.57 & 0.85\\  
        \midrule  
        Macro average & 0.61 & 0.87 \\
        \bottomrule
    \end{tabular}
    \caption{Average Pairwise RMSE (APRMSE) and Spearman correlation coefficients ($\rho$) of 108 random samples of five GPT-4 runs. All Spearman correlations were statistically significant at the level of $p<0.001$.} 
    \label{tab:t1}
\end{table}
Based on these results, we conclude that GPT-4 is reliable in generating appraisal ratings, showing statistically significant Spearman correlations and average pairwise RMSE scores falling into a small tight range. 

\begin{table*}[!ht]
    \centering
    \small
    \setlength{\tabcolsep}{2.7pt}
    \begin{tabular}{lcc|ccccc|c} 
        \toprule
        & \multicolumn{2}{c|} {Q2 - Accuracy}  &  \multicolumn{5}{c|}{Q3 - Majority Voting and Ties}& \multicolumn{1}{c}{Q4 - Emotion}\\
         \cmidrule(lr){2-3} \cmidrule(lr){4-8} \cmidrule(lr){9-9}
        Appraisal & GPT-4& Human$_{\text{avg}}$ & GPT-4$_{\text{avg}}$& GPT-4$_{\text{maj}}$& Human$_{\text{maj}}$ & GPT-4$_{\text{conf}}$& Human$_{\text{conf}}$ & GPT-4$_{\text{emo}}$\\ \hline
        Suddenness & 1.50 & 1.58 &  1.58 & 1.21 & 1.03 & 1.25 & 1.08 & 1.37 \\   
        \underline{Familiarity} & 1.51 & 1.58 & 1.96 & 1.53 & 1.04 & 1.61 & 1.16 & 1.31 \\ 
        Event predictability & 1.44 & 1.56 &  1.70 & 1.31 & 1.08 & 1.38 & 1.16 & 1.14 \\ 
        \bf Pleasantness & 1.11 & 1.23 & 1.15 & 0.59 & 0.57 & 0.61 & 0.63 & 0.59\\ 
        \bf Unpleasantness & 1.20 & 1.33 & 1.32 & 0.73 & 0.68 & 0.72 & 0.71 & 0.69 \\ 
        Goal Relevance & 1.54 & 1.63 &  1.61 & 1.19 & 1.08 & 1.22 & 1.14 & 1.20 \\ 
        Situational Responsibility & 1.43 & 1.65 &  1.67 & 1.10 & 1.03 & 1.12 & 1.14 & 1.50\\ 
        Own Responsibility & 1.34 & 1.45 &  1.60 & 1.03 & 0.83 & 1.06 & 0.91 & 2.26 \\ 
       \underline{Others' Responsibility} & 1.51 & 1.60 &  1.86 & 1.26 & 1.01 & 1.24 & 1.06 & 1.68 \\ \underline{Anticipated Consequences} & 1.60 & 1.69 &  1.68 & 1.32 & 1.21 & 1.35 & 1.28 & 1.57 \\ 
        \bf Goal Support & 1.32 & 1.50 & 1.48 & 0.96 & 0.90 & 0.96 & 0.95 & 1.07 \\ \underline{Urgency} & 1.56 & 1.76 &  1.75 & 1.33 & 1.28 & 1.32 & 1.25 & 1.32 \\ 
        Own Control & 1.44 & 1.56 &  1.43 & 1.04 & 1.05 & 1.12 & 1.16 & 1.06 \\ \underline{Others' Control} & 1.61 & 1.63 &  2.01 & 1.54 & 1.04 & 1.52 & 1.13 & 1.46 \\ 
        Situational Control & 1.58 & 1.65 &  1.66 & 1.10 & 0.99 & 1.12 & 1.14 & 1.10 \\ \underline{Accepted Consequences} & 1.65 & 1.84 &  1.64 & 1.25 & 1.39 & 1.27 & 1.43 & 1.39 \\ 
        Internal Norms & 1.45 & 1.57 &  1.60 & 1.14 & 0.92 & 1.29 & 0.96 & 0.89 \\ 
        \bf External Norms & 1.08 & 1.31 & 1.34 & 0.79 & 0.59 & 0.87 & 0.66 & 0.81 \\ 
        Attention & 1.45 & 1.53 &  1.50 & 1.08 & 1.01 & 1.07 & 1.02 & 1.06 \\ 
        Not Consider & 1.59 & 1.64 &  1.52 & 0.98 & 1.03 & 1.00 & 1.07 & 1.18 \\ 
        Effort & 1.40 & 1.61 & 1.48 & 1.10 & 1.09 & 1.13 & 1.16 & 1.05\\  
        \midrule  
        Macro average & 1.45 & 1.57 &  1.61 & 1.12 & 0.99 & 1.15 & 1.06 & 1.22 \\ 
        \bottomrule
    \end{tabular}
    \caption{RMSE results for research questions Q2, Q3, and Q4. GPT-4 stands for the GPT-4 annotator, Human for the human reader-annotator, \textit{avg}: average of five GPT-4 completions/human guesses, \textit{maj}: majority vote of five GPT-4 completions/human guesses, \textit{conf}: majority vote of five GPT-4 completions/human guesses with confidence rating as a tiebreaker, \textit{emo}: majority vote of five GPT-4 completions with the emotion prediction task in a prompt. Bold/underline marks the appraisal dimensions that are consistently predicted better/worse than the macro average.}
\label{tab:t3}
\end{table*}

\subsection{Q2 Accuracy of GPT-4}
\label{subsec:performance}

After we ensured that GPT-4's responses were reliable on a small subset, we examined how well GPT-4 performed compared to the experiencer-annotators and human reader-annotators on more extensive data. We prompted GPT-4, using the prompt \textit{I1}, to generate appraisal ratings to all 1200 event descriptions in the reader-annotator dataset. Recall, that this data is a subset of the experiencer-annotator dataset, so it has ratings from both the original event experiencer-annotators as well as human reader-annotators.

We calculated the RMSE between the GPT-4 predictions and the experiencer-annotators' responses, which shows how accurately the model can predict the appraisal ratings as a model reader-annotator. We also calculated the RMSE between the experiencer-annotators' responses and human reader-annotators' guesses, which enables to compare how well the model reader-annotator compares to human reader-annotators. Each text in the reader-annotator dataset has five human-reader annotations; therefore, we aggregated them by computing the mean RMSE of five appraisal ratings. 

The results of this experiment are shown in the left-most section of Table~\ref{tab:t3}. We found that the GPT-4 predictions were slightly closer to the human experiencer-annotators' guesses compared to human reader-annotators with an average RMSE over all appraisal dimension of 1.45 compared to 1.57. The results for all distinct appraisals (also in Table~\ref{tab:t3}) reveal that for all appraisal dimensions, the model preditions were closer to human experiencer-annotators compared to human reader-annotators.

We note that the accuracy of the GPT-4-generated appraisal ratings is in the same range to the results reported by \citet{troiano-etal-2023-dimensional} obtained with a fine-tuned RoBERTa model (RMSE = 1.40), with a difference that our results in this section are obtained from just one prediction, whereas the results reported by \citet{troiano-etal-2023-dimensional} are the average over five runs.  Thus, we conclude that GPT-4 is an effective tool for predicting appraisal ratings, performing very close or even slightly better than human reader-annotators.

\subsection{Q3 Effect of Majority Voting}
\label{subsec:majority}

In previous work \citep{troiano-etal-2023-dimensional}, the accuracy of human reader-annotators guesses was considerably improved when their ratings were aggregated using the majority voting over the guesses of five annotators. Encouraged by this, we used GPT-4 to generate five completions for each text in the reader-annotator dataset, and tested whether the majority voting has an impact on prediction accuracy. 
If there was a clear majority vote, that rating was selected as the final prediction. 
In case of ties (e.g., either two ratings were predicted twice or all five ratings were different), we adopted the following essentially random procedure. While each text had five independent predictions from the GPT-4 model, we pretended to have done just a single completion and chose the rating from the first of the five runs.
In total, we generated predictions for 1200 $\times$ 5 data samples using a prompt \textit{I1} with added question "How confident is the model about chosen ratings?".

The application of the majority voting algorithm improved the RMSE on average by about 30\% from 1.61 to 1.12 (columns GPT-4$_\text{avg}$ and GPT-4$_\text{maj}$ in Table~\ref{tab:t3}).  However, compared to the results shown in previous Section~\ref{subsec:performance} (also shown in Table~\ref{tab:t3}), generating five runs and taking the average increased the RMSE considerably, which is now in the same range to the average over the human reader-annotator ratings (1.61 vs 1.57; columns GPT-4$_\text{avg}$ and Human$_\text{avg}$ in Table~\ref{tab:t3}).
At the same time, the RMSE obtained using the majority voting is also considerably lower than the results shown in the previous subsection from just one run (1.12 vs 1.45; columns GPT-4$_\text{maj}$ and GPT-4 in Table~\ref{tab:t3}). Similar or even better results were obtained by applying majority voting to the guesses of the human reader-annotators, showing an improvement from 1.57 to 0.99 (columns Human$_\text{avg}$ and Human$_\text{maj}$ in Table~\ref{tab:t3}). 

Next, we analyzed how often a tie needs to be resolved in the ratings of human reader-annotators and GPT-4. We found that on average, 22\% of human reader-annotators guesses required tie-breaking, while for GPT-4, only 9\% required that. Since the number of ties was not negligible, we considered using a method similar to the one adopted by \citet{troiano-etal-2023-dimensional} based on rating the confidence. Specifically, we added to the prompt's questions list the question ''How confident is the model about chosen ratings?'' as an alternative to the question posed to the human reader-annotator's by \citet{troiano-etal-2023-dimensional} "How confident are you about your answer?''.\footnote{``Answer'' refers to ``emotion'' used in the questionnaire} Similar to other questions, this question was rated on the 5-point Likert scale with values ranging from 1-5. To break a tie, the value with a highest average confidence rating was chosen. 

The results of this experiment are shown in Table~\ref{tab:t3} columns GPT-4$_\text{conf}$ and Human$_\text{conf}$. We can see that using the confidence rating to break the ties did not improve the RSME neither for the model nor for the human reader-annotators compared to the random choice. The mean RMSE for GPT-4 from 1.12 to 1.15 (columns GPT-4$_\text{maj}$ and GPT-4$_\text{conf}$ in Table~\ref{tab:t3}), and the mean RMSE for human reader-annotators increased from 0.99 to 1.06 (columns Human$_\text{maj}$ and Human$_\text{conf}$ in Table~\ref{tab:t3}). When looking at individual appraisal dimensions, we can see that the RMSE's are in most cases lower in the random tie-breaking setting compared to using the confidence rating, and that applies to both the model and humans. Thus, we conclude that using the model-generated confidence rating is a not a useful cue for breaking the ties when aggregating several ratings via majority voting.

\subsection{Q4 Effect of Adding Emotion Prediction}
\label{subsec:emotion}
To study Q4, we again generated five appraisal ratings for each of the 1200 event description in the reader-annotator dataset, using the prompt \textit{I2} that added the task of identifying the target emotion. That means, the prompt asked to pick one emotion from a set of given list of 13 emotion categories (including the no emotion category) most likely felt by the author of the text. This addition made the generation process more error-prone, requiring significantly more runs for entries with shorter descriptions. 

In addition, while generating data for Q4, we again generated ratings (confidence and intensity) to be used as tiebreakers, but this time, we prompted exactly the same questions that were used by \citet{troiano-etal-2023-dimensional} in the human reader-annotator questionnaire. Instead of ``How confident is the model about chosen ratings?'' we asked, ``How confident are you about your answer?''\footnote{Referring to the emotion label prediction.} and for the intensity rating, we asked, ``How intense do you think the emotion was?''. However, we got almost identical results to the random choice voting, which leaves open the question of how to effectively break ties.

The results of adding emotion to a prompt are shown in the last column of Table~\ref{tab:t3}. The average RMSE scores for five completions applying the majority voting algorithm was 1.22 (see column GPT-4$_\text{emo}$), which is worse than the same task without the emotion label prediction studied in relation to research question Q3. 
Thus, we conclude that simply adding the target emotion identification task does not improve the appraisal rating predictions. We note also that the emotion prediction accuracy was ca 55\%, and in 24 cases, GPT-4 generated emotions not present in the predefined list.\footnote{anxiety, disappointment, embarrassment, frustration, jealousy, betrayal, pain, distraction, and indifference} 

\subsection{Q5 Impact of Event Description Length}
\label{subsec:length}

From the beginning of the research, we noted that GPT-4 struggled with shorter event descriptions by often producing inconsistent or empty responses, which suggests that the predictions can be less accurate for shorter texts. Therefore, we analyzed the correlations between event description length (in characters) and the RMSE of the appraisal predictions and examined if the correlation pattern is similar between GPT-4 and human reader-annotators.

We split the reader-annotator dataset into ten bins, each containing even descriptions grouped by length of 100 character intervals 
and compared the average RMSE scores of each bin for both GPT-4 predictions and human reader-annotator guesses. For that experiment, we used that GPT-4 predictions obtained for Q2 in Section~\ref{subsec:performance}.

\begin{figure}[t]
    \centering
    \includegraphics[width=1\linewidth]{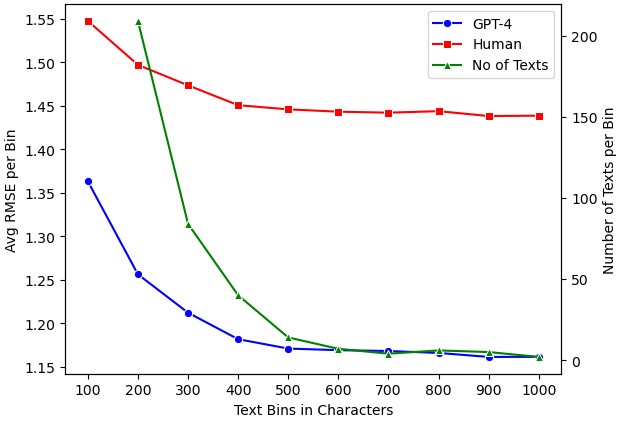}
    \caption{Average RMSE of texts with different lengths. The x-axis labels show the end of the bin in characters: the first bin contains texts with length up to 100 characters, the second between 100 and 200 characters, etc. The secondary y-axis plots the number of texts in each bin, except for the first bin, where the number of texts was ca 800 and thus did not fit to the plot.}
    \label{fig:f1}
\end{figure}

We calculated Spearman's correlation coefficients between the text length and the macro-averaged RMSE for both the model predictions and human reader-annotator guesses. The correlations were negative and very close for the model and humans on average:  $-0.79$ for GPT-4 and $-0.75$ for human reader-annotators, indicating that for both the model and humans, it is harder to predict appraisal ratings for shorter texts. The relation between the text length and the RMSE plotted in Figure~\ref{fig:f1} show that while the human annotator scores show consistently higher RMSE values, the shape of both  curves is similar.

We also looked at the correlations between the text length and the RMSE of each appraisal dimension separately. Two appraisal dimensions indicating the valence of the event (Pleasantness and Unpleasantness) showed near perfect correlations for both GPT-4 and human annotators. Few other dimensions (Suddenness, Own responsibility, Others' responsibility, Situational control) also showed very strong correlations (in the range of $-0.99$ -- $-0.80$). Most other dimensions fell into the strong correlation range ($-0.79$ -- $-0.60$), but some were moderate ($-0.59$ -- $-0.40$).

We also observed that GPT-4 and human annotators had moderate coefficients for different appraisals: Familiarity, Anticipated Consequences, and Not Consider for GPT-4, and Goal Relevance, Urgency, Internal and External Norms for human reader-annotators. 
Finally we note that, with few exceptions, the GPT-4 correlations for individual appraisals tend to be stronger than for human annotators, which means that the accuracy of the GPT-4 predictions is somewhat more dependent on the text length.

Overall, we conclude that there is, on average, a strong negative correlation between the event description length and the accuracy of predicting appraisal rating for both GPT-4 and human reader-annotators. The sweet spot seems to be around 400-500 characters where the RMSE starts to plateau for both GPT-4 and humans.

\section{Discussion}
\label{sec:discussion}

In analyzing appraisal dimensions across all experiments, we looked for patterns by comparing the RMSE of individual appraisal dimensions to the macro-averaged RMSE.
Appraisals consistently showing better accuracy across both GPT-4 models and human annotators include Pleasantness, Unpleasantness, Goal Support, and External Norms (marked as bold in Table~\ref{tab:t3}), suggesting that these are the easiest to infer based on text.
 Own Responsibility, Own Control, Not Consider, and Effort also generally performed well, although Own Responsibility is predicted remarkably worse in the GPT-4$_\text{emo}$ setting.
In contrast, dimensions such as Familiarity, Others’ Responsibility, Anticipated Consequences, Urgency, Others’ Control, and Accepted Consequences consistently show RMSE values equal to or higher than the macro average for both GPT-4 models and human reader-annotators (marked with underscore in Table~\ref{tab:t3}), meaning that the rating of these dimensions were more difficult to infer accurately. 


In contrast to our expectations, asking the GPT-4 model to predict the emotion first and then the appraisal ratings (Q4) did not improve the overall prediction, although the emotion label was expected give useful information about the appraisal values. 
This might be due to the accuracy of the emotion prediction task being moderate, only at 55\%, and incorrect emotion category predictions may have confused the model. 
To check that, we split the data into two groups by correctly and incorrectly predicted emotion labels. Although texts with correctly predicted labels demonstrated better macro average RMSE compared to the texts with incorrect labels (1.21 vs 1.24 RMSE), the difference is not large. However, a closer look at individual appraisal dimensions revealed that most appraisals showed significantly better or comparable accuracy for correctly predicted emotions, with few exceptions of Situational Responsibility (1.57 vs 1.42), Others’ Responsibility (1.70 vs 1.65), Anticipated Consequences (1.65 vs 1.48), and Not Consider (1.25 vs 1.10).  A more fine-grained analysis of why the predictions of these dimensions were more accurate with incorrect emotion labels remains for future research.



We also found that shorter event descriptions generally exhibit lower RMSE values. This result is expected as very short texts convey too little information to predict various appraisal aspects accurately. We found that the optimal event description length that appears to start from the range of roughly 400 to 500 characters and although the improvement in prediction accuracy is more steep for the GPT-4 model, a similar pattern can be observed also for human-readers. This finding has implications for collecting data for emotion appraisal research, as it suggests that researchers should aim for eliciting event descriptions at least 500 characters long.
Finally, throughout the research, we observed that adding more complexity to a prompt resulted in less consistent responses as well as higher RMSE scores. This observation is in line with the reports of other researchers \citep{herderich2024measuring}.

We conclude that GPT-4 is an effective tool for annotating appraisals, though the reliability and accuracy vary across different appraisal dimensions. Our expenses of generating more than 14000 data points to test different strategies were around €200, significantly lower than the £2188 reported by \citet{troiano-etal-2023-dimensional} for 6000 entries annotated by human readers.
Thus, GPT-4 annotations can be a viable alternative to the more costly human reader-annotator ratings in studies requiring large datasets or for generating enough synthetic data for training smaller, local models. 

\section{Conclusions}
In this paper, we studied the reliability and accuracy of GPT-4 in annotating 21 emotion appraisals as an alternative to human reader-annotators. 
The results showed that GPT-4 annotations are highly reliable across several independent runs and it can annotate appraisals with near-human accuracy. Moreover, these results can be considerably improved with using majority voting algorithm over five model completions, which
increased the accuracy of both GPT-4 and human reader-annotators by more than 30\%. Although we tried using predicted confidence rating to resolve the ties, it did not lead to lower RMSE. Thus, there is room for further improvements by finding a better way to resolve the ties.

\section*{Impact}
\label{sec:impact}

In our research, GPT-4 predictions performed similarly to human reader-annotators in annotating appraisal ratings and thus could be applied in psychological research and practice. Predicting user appraisal profile of emotional events could help to identify behavioral and emotional patterns, support therapeutic interventions, or have other practical applications. 

Our study also contributes to the set of effective strategies in predicting appraisal ratings by GPT-4 or potentially similar LLMs.
We show that adopting majority voting algorithm based on five completions can considerably improve the performance of this subjective task. Moreover, we empirically establish an minimum optimal event description length below which both human readers and GPT-4 model prediction accuracy starts to degrade, thus providing a practical guideline for appraisal researchers interested in using predictive models.

The results of this work can inform further research in developing automated reappraisal self-help systems or similar applications, offering practical tools for emotional reframing and appraisal annotation.
However, in those settings, the requirement to submit sensitive and private user content to the GPT-4 API might not be desirable.
An alternative would be to use GPT-4 predictions to augment the training data to improve the accuracy of smaller, local, models that could be subsequently applied in more sensitive data settings.

\section*{Limitations}
\label{sec:limitations}

An important limitation of our research is the limited generalizability of the results. We used only one model, GPT-4, and thus our results cannot be generalized to other available LLMs, although we anticipate that open-source models would probably show poorer results. We also used the specific default configuration of the GPT-4 model. LLMs are sensitive to hyperparameters tuning, and our findings can be applicable only to the used settings. 

Another significant limitation arises from the dataset characteristics. The dataset used in this study reflects limited demographic and socioeconomic diversity, and the emotional events reported by crowd-source participants may be synthetic, potentially reducing their relevance to real-world contexts. 

Finally, GPT-4 and other LLM models can only act as reader-annotators and are limited in the experiencer-annotators roles. 
Thus, our results might have limited impact in psychological research aiming to study the role of appraisals in emotional experiences, as this subjective information can only be provided by human experiencers.

\section*{Ethical Considerations}
\label{sec:ethical}

Using GPT-4 in a reader-annotator context raises ethical questions regarding the accuracy and bias of the annotations. Therefore, GPT-4 annotations should always be validated by both actual event experiencers and human reader-annotators.

Another ethical aspect that has to be taken into account is the privacy and sensitivity of the data. Our research used an open and freely available dataset that does not contain sensitive or private content, collected by \citet{troiano-etal-2023-dimensional} for research purposes.

However, the use this dataset raises some considerations regarding its intended scope and limitations. The dataset was crowd-sourced and as such, we can assume that the participants were a sample from a generally healthy population.
This constraint needs to be kept in mind when using the results of our research in designing systems for clinical domains, such as self-help systems intended to aid in emotional reappraisal for people with clinical issues.

\bibliography{custom,anthology}

\end{document}